\documentclass[final]{article}

\usepackage{compositional_learning}

\usepackage[utf8]{inputenc} 
\usepackage[T1]{fontenc}    
\usepackage{hyperref}       
\usepackage{url}            
\usepackage{booktabs}       
\usepackage{nicefrac}       
\usepackage{microtype}      
\usepackage{xcolor}         
\usepackage{graphicx}%
\usepackage{multirow}%
\usepackage{amsmath,amssymb,amsfonts}%
\usepackage{amsthm}%
\usepackage{mathrsfs}%
\usepackage[title]{appendix}%
\usepackage{color}
\usepackage{xcolor}%
\usepackage{soul}
\usepackage{textcomp}%
\usepackage{manyfoot}%
\usepackage{booktabs}%
\usepackage{algorithm}%
\usepackage{algorithmicx}%
\usepackage{algpseudocode}%
\usepackage{listings}%

\definecolor{lightgrey}{rgb}{0.925, 0.925, 0.925}
\definecolor{orange}{rgb}{0.925, 0.5, 0.0}

\sethlcolor{lightgrey}

\newcommand{\comb}[1]{{\small\texttt{\hl{[#1]}}}}

\title{Successes and Limitations of Object-centric Models at Compositional Generalisation}

\author{%
    Milton L. Montero \\
    Digital Design Department \\
    IT University of Copenhagen \\
    Copenhagen, Denmark \\
    \texttt{mlle@itu.dk} \\
    \And
    Jeffrey S. Bowers \\
    School of Psychological Sciecne \\
    University of Bristol \\
    Bristol, UK \\
    \texttt{j.bowers@bristol.ac.uk} \\
    \And
    Gaurav Malhotra \\
    Department of Psychology \\
    SUNY Albany \\
    Albany, NY, USA \\
    \texttt{gmalhotra@albany.edu}
}

\begin{document}

\maketitle

\begin{abstract}
In recent years, it has been shown empirically that standard disentangled latent variable models do not support robust compositional learning in the visual domain. Indeed, in spite of being designed with the goal of factorising datasets into their constituent factors of variations, disentangled models show extremely limited compositional generalisation capabilities. On the other hand, object-centric architectures have shown promising compositional skills, albeit these have 1) not been extensively tested and 2) experiments have been limited to scene composition — where models  \emph{must generalise to novel combinations of objects in a visual scene instead of novel combinations of object properties}.  In this work, we show that these compositional generalisation skills extend to this later setting. Furthermore, we present evidence pointing to the source of these skills and how they can be improved through careful training. Finally, we point to one important limitation that still exists which suggests new directions of research.

\end{abstract}

\section{Introduction}

A hallmark of human intelligence is compositional generalisation, namely, our ability to perceive and comprehend novel combinations of familiar elements.  For example, in the domain of vision, as long as we can perceive red triangles and blue squares, then we can also perceive blue triangles and green squares.  This gives humans the ability to make “infinite use of finite means” \citep{von1999humboldt,chomsky2014aspects,smolensky_connectionism_1988,mccoy2021infinite} and it is a key priority for AI to achieve human-like abilities.   However, it has proven a challenge for neural network models of vision.

In the context of generative vision models, it was first hypothesized that disentangled representations based on the Variational Auto-Encoder (VAE) architecture could support such compositional generalisation abilities \citep{duan_unsupervised_2020}. While this was a reasonable hypothesis, which had some preliminary experimental support \citep{higgins_beta-vae_2017}, subsequent work showed that such results where not robust and they didn't extend across different levels of difficulty in generalisation \citep{montero2020role,montero2022lost}.Furthermore these results extended to most popular architectures at the time in both supervised and unsupervised settings \citep{schott2021visual}.

Recent work on the other hand has shown that models which perform perceptual grouping — i.e. which decompose images into constituent objects — exhibit increased compositional generalisation capabilities \citep{singh2021slate,frady_learning_2023,wiedemer_compositional_2023}. However, these results typically focused on scene compositions i.e. generalisation to novel configurations of known objects in a scene. This is however, a fundamentally different challenge to the one posed by composition of different object properties, both intrinsic (like shape and color) and extrinsic ones(like position and rotation), since the former requires manipulating the relation between objects and the later the relation between their parts and how this is translated from proximal to distal representations \citep{pizlo2001perception}.

This has left the problem of compositional generalisation across novel combinations of \emph{object properties} relatively unexplored. Instead most research in this area this area is focused on text-to-image (e.g. such as DALL-E) which assume the presence of language, or the use of Geometric Deep Learning approaches which exploit invariances and symmetries in the domain \cite{boresntein2021}.

In this work we turn our attention back to this question of object-level compositional generalisation (which we will refer to as simply compositional generalisation), but instead focus on exploring the capabilities of models that poses less inductive biases than the ones using either language or invariances. We will test object-centric models (specifically SlotAttention \citep{locatello2020objectcentric} and a specific variation which we introduce later) on standard compositional generalisation tasks. Our contribution are as follows:
\begin{enumerate}
    \item We show that SA can solve object-level compositional generalisation conditions that were found to be challenging in \citet{montero2022lost} — specifically, the condition described there as recombination-to-range (R2R), where a range of combinations for two factors are removed from training.
    \item We show that the model still struggles when shape needs to be combined with novel rotations, like due to this condition requiring the model to reconstruct novel local features when shapes are rotated.
    \item We introduce a novel dataset to test whether this theory, and show that when novel rotations only change the global configuration of a shape but not it's local features the models then succeeds.
    \item Finally we show that when trained with this novel dataset, the model even manages to solve extrapolation conditions where it must reconstruct novel shapes.

\end{enumerate}

\section{Object-level Compositional Generalisation in Object-centric Models} \begin{figure}[!t]
    \centering
    \includegraphics[width=0.49\textwidth]{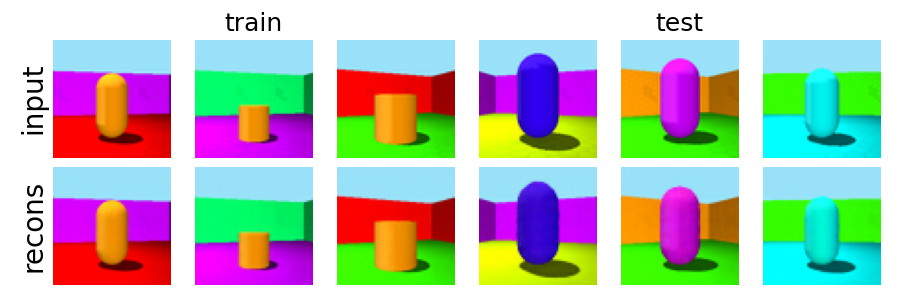};
    \includegraphics[width=0.49\textwidth]{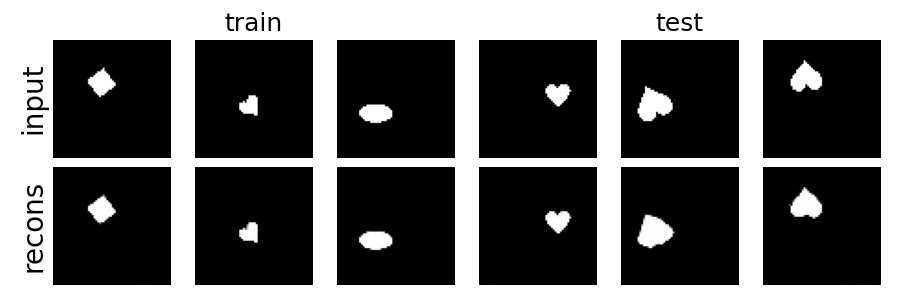}
    \caption{\textbf{Slot Attention generalisation results}: Reconstructions for a model when trained on all but some combinations of generative factors. The model is tested on said excluded combinations. Left) Generalisation results when excluding half of the combinations of the colors with the pill shape in 3DShapes. Right) Analogous test on dSprites where we exclude half of the rotations of the heart.}
    \label{fig:sa-generalisation-results}
    \vspace{-0.5cm}
\end{figure}

It was argued in \citet{montero2022lost} that a possible cause for the failures at combinatorial generalisation described in \citet{montero2020role} and \citet{schott2021visual} was that, despite learning disentangled representations, the generative models did not segment images into their constituent parts. Thus, when faced with novel combinations of properties that determine the same element of an image (e.g. novel combinations of shape and color), they could not manipulate the individual elements in an image in order to change the relevant property (e.g. the location) without affecting the representation of other properties.

\begin{table}[h!]
    \centering
    \caption{\textbf{Quantitative results on standard datasets.} Pixel-wise sum of squared errors for both an SA and a WAE model on the same datasets and conditions showed in Figure~\ref{fig:sa-generalisation-results}. For 3DShapes models with a score below 20 tend to be visually indistinguishable. For dSprites, the same effect tends to happen at 10.}
    \begin{tabular}{lllll}
    \toprule
    & \multicolumn{2}{c}{Shapes3D} & \multicolumn{2}{c}{dSprites}  \\
    Model & Train & Test & Train & Test \\
    \midrule
    SlotAE & 1.63 & 9.76 & 1.50 & 3.06 \\
    WAE & 18.66 & 180.34 & 7.10 & 14.20  \\
    \bottomrule
    \end{tabular}
    \label{tab:sa-vs-wae_dsprites_and_shapes3d}
\end{table}

We test a SlotAttention (SA) model on these same conditions for which disentangled models failed in \cite{montero2022lost}: novel combinations of shape and position in dSprites, and novel combinations of shape and color in 3DShapes (see Appendix~\ref{app:datasets} for more details). Results can be seen in Figure~\ref{fig:sa-generalisation-results}.

The results on the left show a clear success for the 3DShapes dataset, where previous disentangled models failed catastrophically. Given the architectural properties of SA it is also easy to pinpoint to why this can be the case. First, SA possesses in-built positional knowledge in the form of position embeddings. This allows the model to capture the relation between different patches in the scene, and amongst them, the ones that encode parts of the same object. This should allow the model to map how those different patches will change when asked to produce a different rotation. Second, the SlotDecoder forces reconstructions to be performed on a per-object basis, reinforcing the innate bias of the model to limit it's representations to aggregations of patches that should be manipulated together.

The above is further illustrated with the results obtained on dSprites. On the right, the model fails when tested on novel combinations of shape and rotations on dSprites (half of the rotations of the heart where excluded). However, while the model reconstructions are worse, they do not constitute a complete failure as was the case in previous studies. In fact it seems like the model correctly identifies the required shape and rotation, but fails to properly reconstruct them. We hypothesize that in this image space, features are orientation specific which means that removing some of them prevents the model from learning how to reconstruct them. In the next section we explore how far we can push these models when we correct for said issue in the training data.

\subsection{Pushing the Generalisation Capabilities of Object-centric Models}

\begin{figure}[t!]
    \centering
    \includegraphics[width=0.7\textwidth]{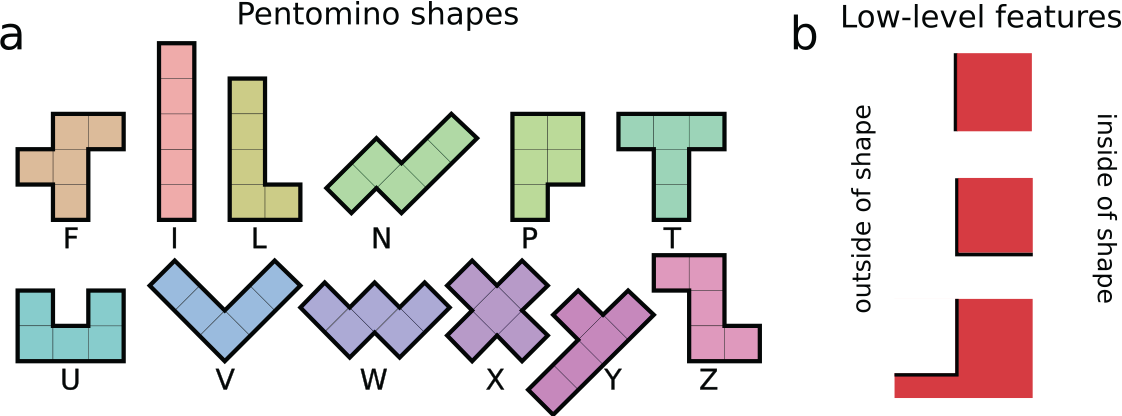}
    \caption{\textbf{Pentomino shapes} a) The twelve Pentomino shapes and their names\protect\footnotemark. We construct the dataset by performing affine transformations of these shapes: 5 values of scale, 40 values of rotation and 20 values of translation along each of the X and Y axis. b) The low-level features that comprise the different shapes. From top to bottom: straight lines, convex right angles and concave right angles.}
    \label{fig:pentominos-small}
    \vspace{-0.5cm}
\end{figure}

\footnotetext{Modified from \href{https://commons.wikimedia.org/wiki/File:Pentomino_Naming_Conventions.svg}{Pentomino Naming Conventions}. \href{https://commons.wikimedia.org/wiki/User:Nonenmac}{R.A. Nonenmacher}, \href{https://creativecommons.org/licenses/by-sa/4.0}{CC BY-SA 4.0}, via Wikimedia Commons.}

To test whether the previous failures when known shapes are presented in novel rotations are the result of models not having access to the correct local features during learning, we create a dataset where all local features are trained and test whether the model succeeds when tested on equivalent conditions (excluded shape and rotation combinations). We designed said dataset using the Pentomino shapes: sets of five blocks arranged into different shapes which we vary along different factors of variation as in dSprites ( (Figure~\ref{fig:pentominos-small}), see Appendix~\ref{app:pentomino-data} for more details).

Notice that in this dataset all low-level features are straight lines or right angles (Figure~\ref{fig:pentominos}, panel b). Thus even when presented with a novel combination of shape and rotation as before, we can be more confident that the low-level features are not novel and the model only needs to respect the global configuration of the different parts of the shape. This is in contrast to dSprites where a novel rotation of a shape such as the heart requires reconstructing a novel local-feature (e.g. the small dip of the heart in a novel rotation is effectively a novel feature in the image frame of reference).

To facilitate our analysis we will test a simplified version of the SA architecture. In this version there is only one slot which must perform figure ground segmentation (hence we name it FgSeg). As in SA the model, it uses an attention mechanism, but instead of slots competing to explain patches of data, the latent must only integrate information about the foreground while ignoring the background. During reconstruction we use a simplified version of the SlotDecoder where instead of using a Softmax to decide which slot is responsible for reconstructing a particular pixel, we use a Sigmoid activation function which determines if the latent must reconstruct a particular pixel. Thus the model is encouraged to only represent the Pentomino shape, and not the full image. This makes analysing the model easier as having slots compete to explain the Pentomino object (a.k.a. the figure) tends to split said object amongst the different slots. Additionally, it is easier to explore how the object is represented if we know it is encoded in the only available latent.

\subsection{Compositional Generalisation Results}

\begin{figure}[t!]
    \centering
    \includegraphics[width=\textwidth]{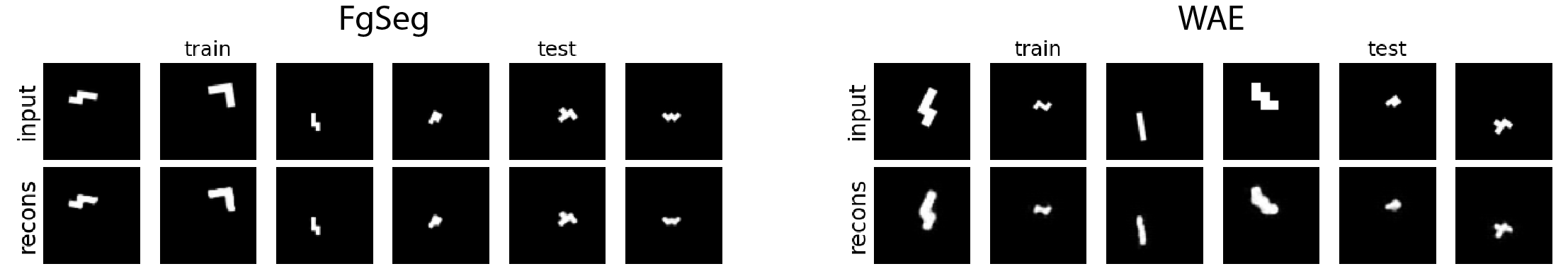};
    \caption{\textbf{Generalization to novel shape and rotation combinations in the Pentomino dataset}. Generalization reconstructions for both FgSeg and a WAE control model. The models where trained on 11 of the 12 Pentomino shapes and tested at reconstructing a held out one in different configurations of position, rotation and scale.}
    \label{fig:pentomino-rot}
    \vspace{-0.5cm}
\end{figure}

We train FgSeg on the Pentomino dataset, excluding half of the rotations for a 4 of the shapes (out of a total of 12). We then test the model on these excluded shapes as before. To control that the model performance is not only due to the qualitative differences in the dataset, we compare against a Wasserstein Auto-Encoder (WAE, \citet{tolstikhin2017wasserstein}) tested on the same generalisation condition (Figure~\ref{fig:pentomino-rot}).

The figure clearly shows that, while the WAE fails to reconstruct novel rotations of a known shape, the FgSeg model succeeds, which shows that a perceptual grouping model can solve previously challenging generalisation conditions given an appropriate dataset — one where novel combinations of factors do not imply the presence of novel local features in image space. 

A quantitative measure is presented in Table~\ref{tab:sa-vs-pentomino-shape-and-rotation}, showing that the qualitative examination is supported by the scores achieved by the model. In this case, training scores refers to validation on a randomly sampled held-out dataset. In all cases the FgSeg model achieves better scores than the WAE model, though the generalisation for property prediction, while better, still doesn't match the validation performance.

\begin{table}[h!]
    \centering
    \caption{\textbf{Novel shape-rotation combination scores on the Pentomino dataset.} Scores for both FgSeg and the baseline WAE on the Pentomino dataset. Reconstruction scores are in P-MSE, while rotation prediction uses plain MSE. Classification is in accuracy.}
    \begin{tabular}{lllllll}
    \toprule
    & \multicolumn{2}{c}{Reconstruction} & \multicolumn{2}{c}{Shape Classification} & \multicolumn{2}{c}{Rotation Prediction} \\
    Model & Train & Test & Train & Test & Train & Test \\
    \midrule
    WAE & 5.30 & 10.55 & 48.97 & 25.76 & 0.20 & 0.48 \\
    FgSeg & 1.11 & 2.15 & 97.7 & 49.92 & 0.022 & 0.42 \\
    \bottomrule
    \end{tabular}
    \label{tab:sa-vs-pentomino-shape-and-rotation}
\end{table}

\subsection{Extrapolation results}

\begin{figure}[h!]
    \centering
    \includegraphics[width=\textwidth]{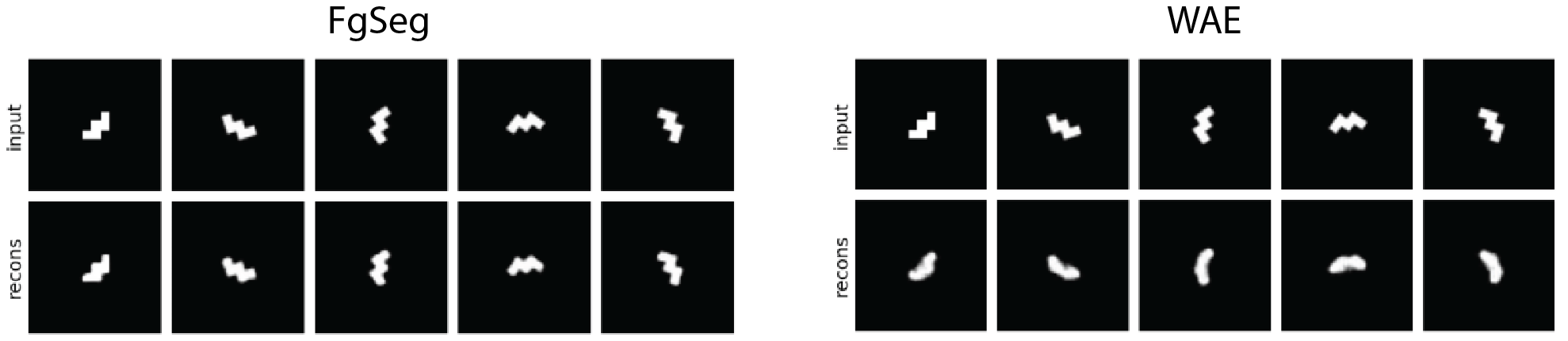}
    \caption{\textbf{New shape extrapolation} On the left, Slot Attention reconstructions of a novel shape, in this case the W. Left to right, different values of rotation sampled uniformly over the whole range of values $[0, 360)$ can be seen. On the right, the same results for WAE. It is clear that SA succeeds where WAE does not.}
    \label{fig:pentomino-extrapolation}
    \vspace{-0.5cm}
\end{figure}

Given this success, what happens if we now test the model on an extrapolation condition — where the model must reconstruct completely novel shapes? We show that, perhaps surprisingly, the model can also succeed when tested on reconstructing three novel shapes in the Pentomino dataset, effectively learning how to recombine the local features in the Pentomino dataset into potentially arbitrary shapes (Figure~\ref{fig:pentomino-extrapolation}).

We quantify this success, showing how the FgSeg model's reconstruction scores change as we exclude more and more shapes from the training data. We see that between that there is no significant drop from 1 to three, and only after removing half of the shapes (6) do we start to see a drop in performance.

\begin{table}[h!]
    \centering
    \caption{\textbf{Pentomino shape-rotation generalisation.} Reconstruction scores for extrapolation conditions of increasing number of novel shapes. Reconstruction error in P-MSE.}
    \begin{tabular}{llll}
    \toprule
    Data Split  & One Novel Shape & Three Novel Shapes & Six Novel Shapes \\
    \midrule
    Train & 1.65 & 1.29 & 1.25 \\
    Test & 2.47 & 2.63 & 6.37 \\
    \bottomrule
    \end{tabular}
    \label{tab:pentomino-extrap-scores}
\end{table}

\section{Discussion}

We have shown that SA can solve compositional generalisation challenges that posed a significant issue for standard auto-encoder generative models. Furthermore, we have shown that failures in those datasets are likely due to the fact that some local features are effectively removed from the dataset when we exclude certain combinations, and that if we control for this fact, we can achieve generalsation for more complex factor combinations such as novel combinations of shape and rotation. 

To our surprise, when trained and tested on this qualitatively different dataset, the model was even able to solve extrapolation tasks that require reconstruction of unseen shapes. This shows once a again that careful data curation can have a significant effect on model capabilities when combined with the right architecture. Our results also show that we may not require stronger inductive biases such as the ones introduce in \cite{singh2022neural} to solve these tasks, and that separation into discrete tokens describing values of the properties is likely only needed for higher-level cognitive tasks (such as reasoning and planning)

Indeed our follow-up experiments suggest that the models learn abstract representations of the concepts present in said dataset. When using the learned embeddings from FgSeg to fit a classification model for the shape property, we find that a linear classifier is unable to correctly solve the problem, and instead we must use a non-linear one such as a Support Vector Machine. Even then, such a classifier is unable to correctly classify the held-out combinations from using the test embeddings without jointly training both i.e. it does not generalise in the same way the reconstruction does (see Appendix~\ref{app:predictivity}). Extracting or learning such higher level concepts using/from the representations produced by these simpler models is thus an interesting direction for future research.

One way to accomplish this could be to design architecture that incorporate more principles from the Gestalt theory of perception, of which FgSeg and and SA only implement a subset of, namely figure-ground segmentation and perceptual-grouping \citep{wagemans2012century,yantis2001visual,treisman1980feature}. Principles such as common-fate (the idea that the visual system is biased towards grouping together features that move together in time to the same object representation), could potentially further constrain the model and induce more general representations of shape, and idea that has been preeliminary explored in \citet{tangemann2021unsupervised}.

\acksection
\vspace{-0.4cm}
The authors would like to thank the members of the Mind \& Machine Learning Group for useful comments throughout the different stages of this research. This research was supported by a ERC Advanced Grant, Generalization in Mind and Machine \#741134.


\bibliographystyle{unsrtnat}
\bibliography{bibliography}

\medskip






\newpage

\appendix

\section {Limitations}\label{app:limitations}

We have presented our results using a single model. In our preliminary tests we trained several models on the generalisation conditions and found that there was no significant difference in performance for those baseline tests. As such, we conducted the rest of the experiments using only one seed per tests. Additionally, we also tested SLATE on the same conditions of dDsprites  and 3DShapes and found that it performed qualitatively the same. As such we elected to not test it further as we believe it unlikely that it would perform differently.

Another limitation is that visualising the latent space provided no useful information. This is unsatisfying since raw scores tend to not provide a full picture of how the models are performing a given task.

\section{Datasets}\label{app:datasets}

We first test SA models on hard combinatorial generalisation conditions such as the recombination-to-range ones identified in (Montero et al., 2022). Namely, we test the following conditions for the three datasets:

\begin{itemize}
    \item \textbf{3DShapes}: Contains the 6 generative factors: \comb{floor hue, wall hue, object hue, scale, shape, orientation}. Colors are defined in the HSV format, and the values correspond to the hue component. Here orientation defines the angle of point-of-view for the scene. The objects themselves do not rotate. The hard condition in this case is all images such that \comb{shape=pill, object-hue=$>0.5$}, which were excluded from the training set. Again, these are pills colored as any of the colors in the second half of the HSV spectrum. These colors (shades of blue, purple, etc) were observed on the other shapes, and the pill was observed with other colors such as red and orange. See \citet{3dshapes18}.
    \item \textbf{dSprites}: Contains the following generative factors: \comb{shape, scale, orientation, position X, position Y}. Orientation here refers to the rotation of the shape along it's center of mass. The hard condition is all images such that \comb{shape=heart, rotation$<180$}, which were excluded from the training set. Thus, no squares rotated beyond the 180 are seen during training and the model must reconstruct them at test time. We also excluded redundant rotations for the training data since shapes such as the ellipsis and the square, unlike the heart, are less than $360^\circ$ symmetrical. See \citet{matthey_dsprites_2017}.
\end{itemize}
Notice that excluding a combination of a shape with another factor means that said shape will not be seen the same number of times during training. For example in the dSprites condition defined above squares will be observed half as many times as the other two shapes. To ensure that shapes are observed an equal amount of times, we sample instances from the dataset with the following probabiliy:

\begin{equation}\label{eq:dataset-rebalance}
    p(x_i, g_i)= \frac{1}{|\{(x_j, g_j) \in \mathcal{D}_{train} \mid g_j[\texttt{shape}] == g_i[\texttt{shape}]\}|}
\end{equation}

This ensures that images in the training set that belong to the shape that is used in the generalisation condition (e.g. the square in dSprites) are seen as frequently as other shapes. Of course there will be less variation in the factor that is used to test said generalisation condition (e.g. position along the X-axis), however these have a larger number of values so they are less likely to be the cause of over-fitting.

\subsection{The Pentominos Dataset}\label{app:pentomino-data}

To adjudicate between these two views we unfortunately cannot rely on the dSprites dataset since the shapes used to generate it share few low-level features amongst them. Indeed the most prominent features (the curve of the ellipsis, the right angles of the square and the dip of the heart) are unique to each of them. Thus we introduce a new dataset based on the \hyperlink{https://en.wikipedia.org/wiki/Pentomino}{pentomino shapes} to tackle this issue.

\begin{table}[h!]
\centering
\caption{\textbf{Pentomino generative factors}. The generative factor values used to generate the Pentomino dataset used in our experiments.}
\begin{tabular}{ |p{3cm}|p{3cm}|p{5cm}|  }
\hline
Generative factor & No. of values & Values \\
\hline
Shape      & 12 & F, I, L, N, P, T, U, V, W, X, Y, Z \\
Scale      &  5 & 1.5, 1.8, 2.1, 2.4, 2.7, 3.0 \\
Rotation   & 40 & Evenly spaced in ($0^\circ$, $351^\circ$) \\
Position-X & 20 & Evenly spaced in (-1, 1) \\
Position-Y & 20 & Evenly spaced in (-1, 1) \\
\hline
\end{tabular}
\label{tab:pentomino-gen-fact}
\end{table}

\begin{figure}[t!]
    \centering
    \includegraphics[width=\textwidth]{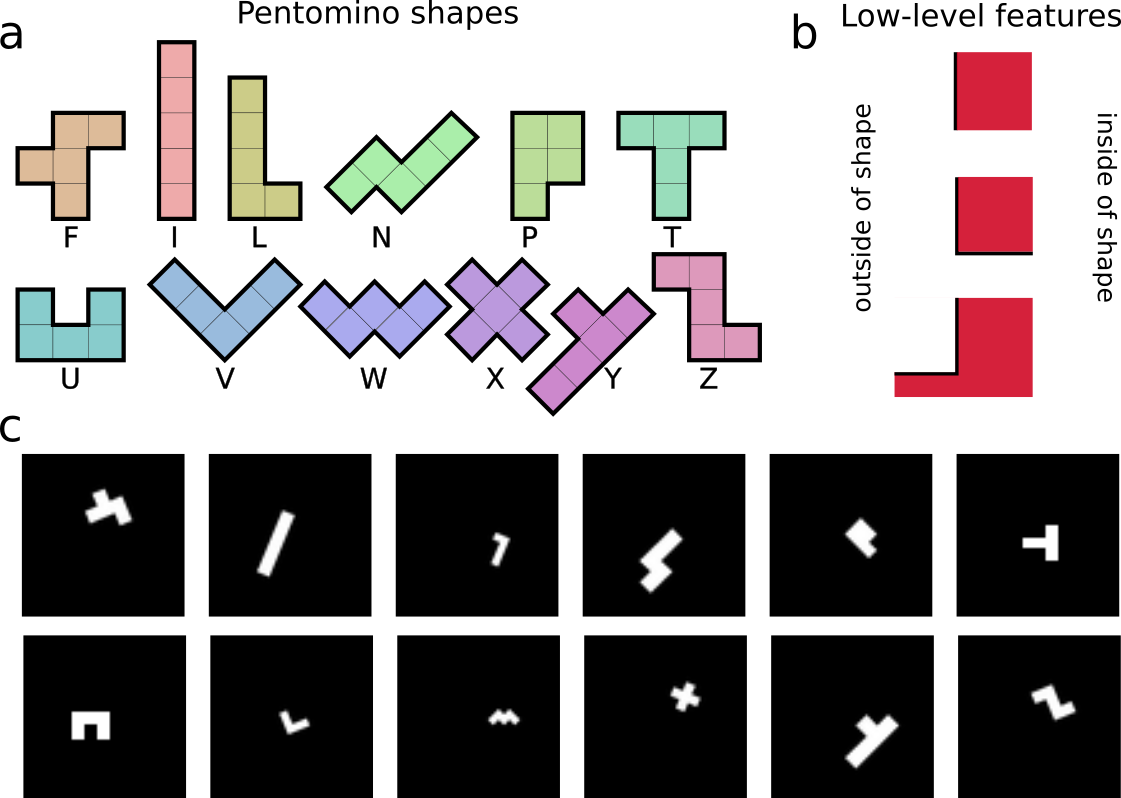}
    \caption{\textbf{Pentomino shapes} a) The twelve pentomino shapes and their names\protect\footnotemark. We construct the dataset by performing affine transformations of these shapes: 5 values of scale, 40 values of rotation and 20 values of translation along each of the X and Y axis. b) The low-level features that comprise the different shapes. From top to bottom: straight lines, convex right angles and concave right angles. c) Example stimuli containing different configurations of the different factors, with each shape represented once.}
    
    \label{fig:pentominos}
\end{figure}

The pentomino shapes are simple, sprite-like shapes, composed of five equal side squares in different configurations that are joined edge-to-edge (see Figure~\ref{fig:pentominos}.a). There are twelve such shapes in total without taking into account rotations or mirror symmetries (the so-called free pentominos). It is then clear how these shapes solve our issue: The low-level features that compose all the shapes are straight-lines, convex right angles and concave right angles (Figure~\ref{fig:pentominos}.b). Moreover, all shapes have at least one instance of each of these features with the sole exception of the ``I'' which does not have a concave right angle (notice that it is impossible to create a polygon without the other two).

To generate the dataset we use similar variations in the generative factors as in dSprites with an important caveat. Because there are 12 shapes as opposed to 3, we reduce the range of values that three of the generative factors (scale, position x, and position y) can take so that the dataset is not much larger. We preserve the 40 values of rotation since this is one of the factors we wish to test. For the rest, apart from the already established 12 values of shape, we use 5 values of scale and 20 values for the position along each of the X and Y axis. This amounts to a total of 960000 (as opposed to the 737280 examples in dSprites). See Table~\ref{tab:pentomino-gen-fact}.

\footnotetext{Modified from \href{https://commons.wikimedia.org/wiki/File:Pentomino_Naming_Conventions.svg}{Pentomino Naming Conventions}. \href{https://commons.wikimedia.org/wiki/User:Nonenmac}{R.A. Nonenmacher}, \href{https://creativecommons.org/licenses/by-sa/4.0}{CC BY-SA 4.0}, via Wikimedia Commons.}

\section{Extra results}
\subsection{Shape-rotation Generalisation}\label{app:shape-rot-results}

As in the previous section, we define a combinatorial generalisation condition that excludes some shape and rotation values:
\begin{itemize}
    \item \textbf{Novel shape and rotation combinations}: We exclude combinations such that \comb{shape $\in \{F, P, T, W\}$, rotation > $180^{\circ}$}. We selected these shapes so that there are a couple of shapes that are similar to other shapes in the training data at those rotations (T and Y, W and V or U), and other two (F and P) that are very distinct. The fist pair tests if the model will confuse when the rotation value is novel and the latter
    if it will be able to produce a reconstruction for which it is harder to interpolate. Four shapes also allows us to maintain a similar ratio of 1:3 excluded to included during training for the shape factor (4 of 12 excluded here vs 1 of 3 in dSprites).
\end{itemize}

As in the previous section, we remove the redundant rotations of the I, X and Z shapes and correct for the unbalance of presentations of the shapes as defined by Equation~\ref{eq:dataset-rebalance}. We use the same architecture and training configuration that we used for dSprites for both FgSeg and WAE. We use the latter as a baseline. We also use this baseline to control for the increase in the amount of shapes with respect to dSprites. If FgSeg was to succeed in this new dataset, it could be argued that this is because having twelve shapes gives the model more opportunity to learn a proper representation of what constitutes a shape. A success at generalisation by FgSeg and failure from a baseline model would rule out this possibility, which means the former's success is much more interesting than if both succeed.

\begin{figure}[t!]
    \centering
    \includegraphics[width=\textwidth]{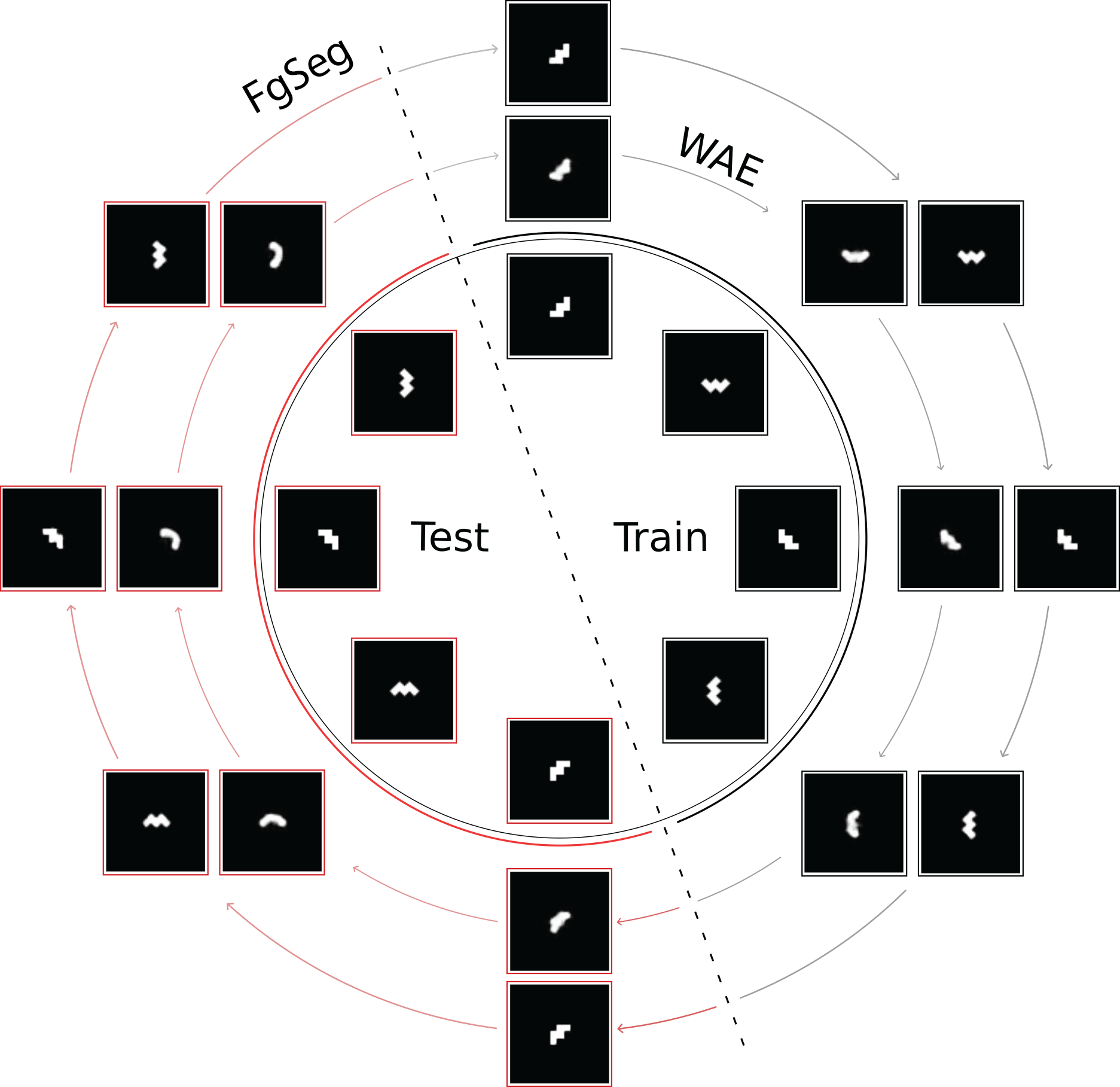}
    \caption{\textbf{Generalization to novel shape and rotation combinations in the Pentomino dataset}. Generalization reconstructions for both FgSeg and a WAE control model. Example rotations of the excluded shape are plotted inside their circumference according to their rotation value. Samples highlighted in red are novel combinations only seen during testing. The rest (highlighted in black are used during training. Reconstructions for FgSeg and WAE are plotted on the outside in the corresponding angle. FgSeg is always outside and WAE inside.}
    \label{fig:pentomino-shape-and-rotation}
\end{figure}
o

The results can be seen in Figure~\ref{fig:pentomino-shape-and-rotation}. Reconstructions are plotted along the circumference according to the ground truth rotation value. For simplicity we keep the other generative factors fixed at the midpoint value. In the case of FgSeg we see that the results are very impressive. The model shows no sign of confusing either to rotation or the shape of unseen combinations. Thus they clearly support the second view over the first one described in the introduction to this section: errors committed on novel shape-rotation combinations are due to decoder errors.

It is also clear that said improvement in generalization is not due to the increase in the number of shapes as the results for WAE show clear and systematic failures at generalisation, especially for rotations that are far from the ones observed during training, as should be expected.

\subsection{Additional Extrapolation Results}\label{app:pentomino-extrapolation}

We also test the model when excluding more than one shape from traning to test the degree to which variety in the training examples influences the quality of the learned representations/genertive mechanism:

\begin{itemize}
    \item \textbf{Three novel shapes}: We exclude all instances where \comb{shape $\in \{P, T, W\}$}. We include P and T because when we tested WAEs these were routinely confused with F and Y respectively. Thus it may increase the likelihood of SA failing by making the same mistakes.
    \item \textbf{Half novel shapes}: We further increase the number of shapes by excluding inputs such that \comb{shape $\in \{F, P, N, T, V, W\}$}. We just add F, V, Z to make it 6 out of 12 shapes excluded.
\end{itemize}

We also including raw reconstruction errors for all three conditions, including the one in the main text (Figure~\ref{fig:pentomino-extrapolation}).

\begin{figure}
    \centering
    \includegraphics[width=\textwidth]{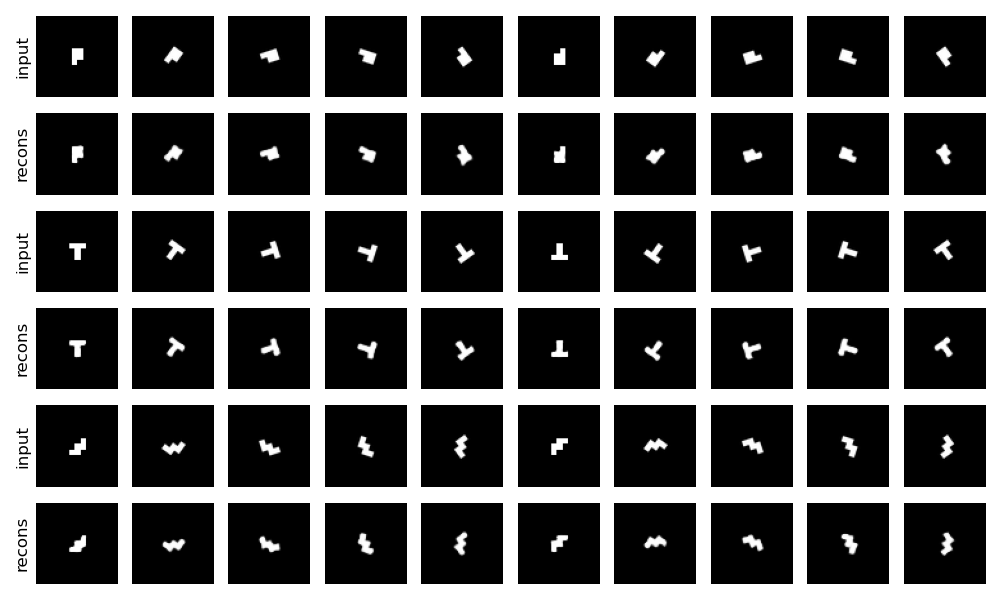} %
    \caption{\textbf{Extrapolation to three new shapes}. Figure-ground Segmentation model reconstructions when three shapes are excluded from training (P, T, W). Every pair of consecutive rows contains images of one of the shapes at 10 different rotation values. These are taken at evenly spaced angels from $180^\circ$ to $360^\circ$ and plotted increasingly from left to right. The rest of the generative factors are kept constant. We can observe that the model achieves good quality reconstructions in spite of not having seen these shapes at all.}
    \label{fig:pentomino-three-new-shapes}
\end{figure}

\begin{figure}
    \centering
   \includegraphics[width=\textwidth]{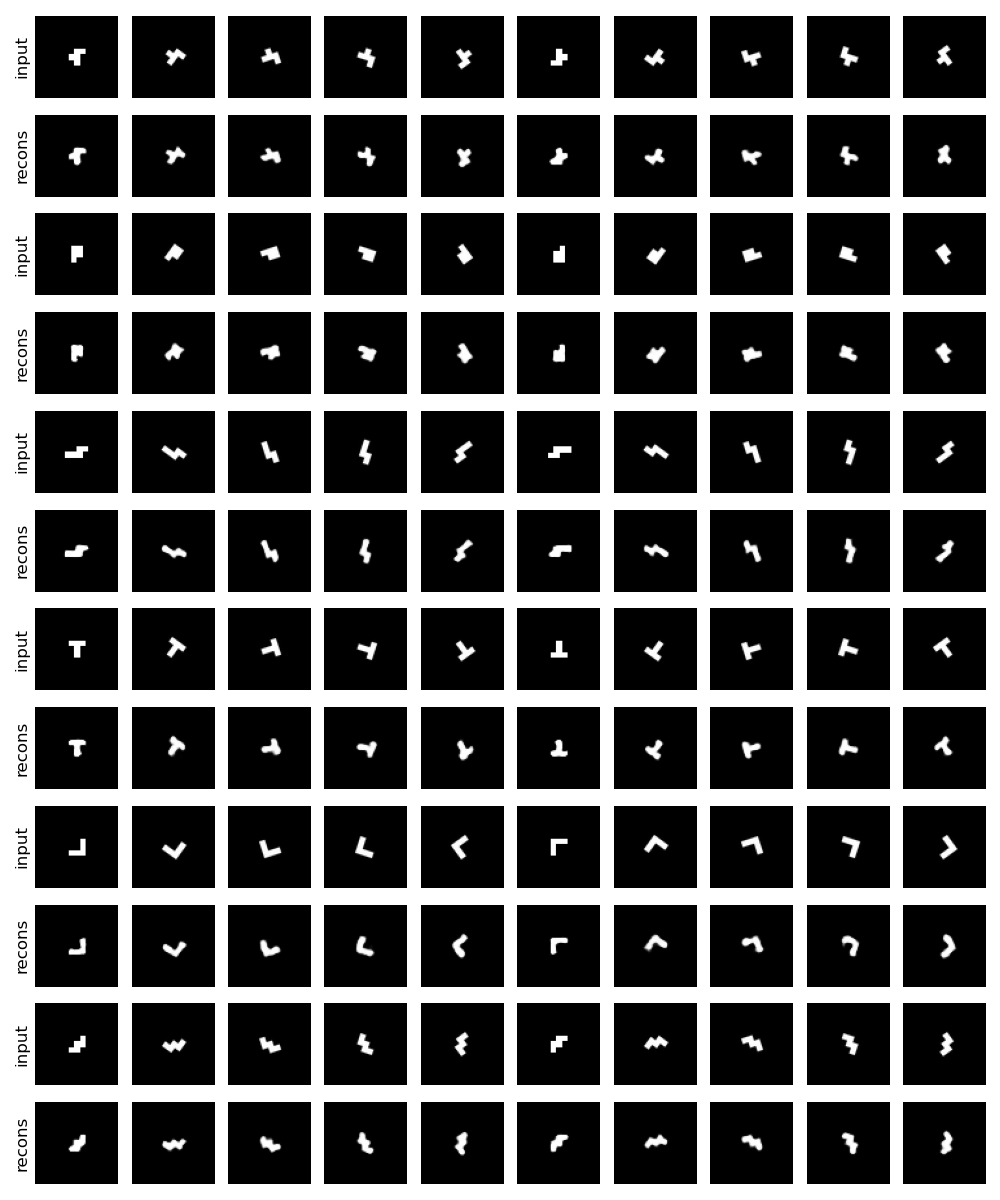} %
    \caption{\textbf{Extrapolation to six new shapes}. Figure-ground Segmentation model reconstructions when six shapes are excluded from training (F, P, N T, V, W). Every pair of consecutive rows contains images of one of the shapes at 10 different rotation values. These are taken at evenly spaced angels from $180^\circ$ to $360^\circ$ and plotted increasingly from left to right. The rest of the generative factors are kept constant. Unlike the previous example, model reconstructions start to show significant degradation.}
    \label{fig:pentomino-half-new-shapes}
\end{figure}

On the other hand, removing more shapes does produce significant degradation in the reconstruction quality. The results for this condition can be observed in Figure~\ref{fig:pentomino-half-new-shapes}. For some shapes such as W we can see that the corners are not as well defined as before. For others like the T, the deformations are more pronounced. In the case of the V it seems to even confuse it with the W on some examples. Nonetheless, these results are still better than what we obtained with a WAE when excluding only one shape. Thus we can conclude that while there is an effect of data diversity, the model is fairly good at generalization even when a large number of examples are excluded form the training data.

\section{Testing predictivity of learned representations}\label{app:predictivity}

Are the learned representations high level? In other words does the model learn concepts related to each shape and color or does it solve the task using simpler, lower level representations? Unlike disentangled VAEs it is not easy to directly visualise the latent representations of these models. Instead we turn to prediction of the different concepts (such as the shape classes) as a way to assess this. If models learn representations that are abstract, we should be able to use them to predict the classes for both training and held out samples. We test this below. First of all, we see that models do learn representations that are decodable using linear probes since we need at least a two-layer MLP to achieve 100\% accuracy on the training data. An even these models cannot achieve significant prediction accuracy in the test data, which shows that the model's representations are not abstract after all.

\begin{figure}
    \centering
    \includegraphics[width=\textwidth]{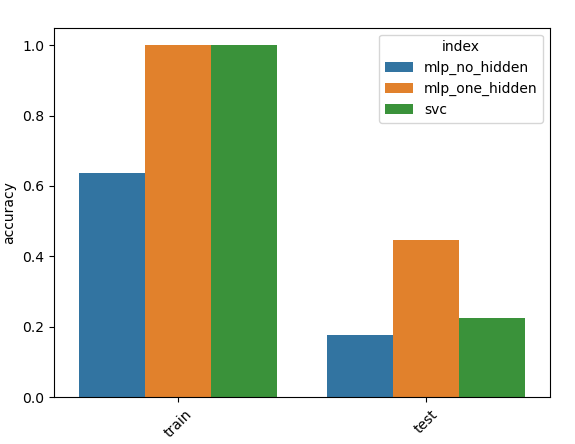} %
    \caption{\textbf{Testing abstract representations}. On the left accuracy of three different probing models when prediction the shape of training images using the representations learned by a Slot Attention model: A simple Linear classifier trained with SGD, an MLP with one hidden layer and a Support Vector Machine. On the right, the same models now tested on prediction shapes for novel images using the slot embeddings obtained from the model.}
    \label{fig:prediction}
\end{figure}

\end{document}